\documentclass[conference]{IEEEtran}
\usepackage{microtype}
\usepackage{graphicx}
\usepackage{subfigure}
\usepackage{booktabs} 
\usepackage{amsmath}
\usepackage{bbm}
\usepackage{hyperref}
\allowdisplaybreaks[1]

\begin{document}

%
\title{GPM: A Generic Probabilistic Model to Recover Annotator's Behavior and Ground Truth Labeling}
\makeatletter
\newcommand{\linebreakand}{%
  \end{@IEEEauthorhalign}
  \hfill\mbox{}\par
  \mbox{}\hfill\begin{@IEEEauthorhalign}
}
\makeatother

\author{

\IEEEauthorblockN{ Jing Li }
\IEEEauthorblockA{\textit{MOKU Lab, Alibaba Group} \\
jing.li.univ@gmail.com}
\and
\IEEEauthorblockN{ Suiyi Ling}
\IEEEauthorblockA{\textit{Capacit\'{e}s SAS} \\
suiyi.ling@univ-nantes.fr}
\and

\IEEEauthorblockN{ Junle Wang}
\IEEEauthorblockA{\textit{Turing Lab, Tencent} \\
wangjunle@gmail.com}

\linebreakand
\IEEEauthorblockN{ Zhi Li}
\IEEEauthorblockA{\textit{Netflix} \\
zli@netflix.com}
\and
\IEEEauthorblockN{ Patrick Le Callet}
\IEEEauthorblockA{\textit{IPI/LS2N Lab, University of Nantes} \\
patrick.lecallet@univ-nantes.fr}

}

%

%

\markboth{Journal of \LaTeX\ Class Files,~Vol.~xx, No.~x, xxxx~2020}%
{Shell \MakeLowercase{\textit{et al.}}: Annotation Model}

\maketitle
\begin{abstract}
In the big data era, data labeling can be obtained through crowdsourcing. Nevertheless, the obtained labels are generally noisy, unreliable or even adversarial. In this paper, we propose a probabilistic graphical annotation model to infer the underlying ground truth and annotator's behavior. To accommodate both discrete and continuous application scenarios (e.g., classifying scenes vs. rating videos on a Likert scale), the underlying ground truth is considered following a distribution rather than a single value. In this way, the reliable but potentially divergent opinions from "good" annotators can be recovered. The proposed model is able to identify whether an annotator has worked diligently towards the task during the labeling procedure, which could be used for further selection of qualified annotators. Our model has been tested on both simulated data and real-world data, where it always shows superior performance than the other state-of-the-art models in terms of accuracy and robustness.
\end{abstract}

%
%

%

\IEEEpeerreviewmaketitle

\section{Introduction}
\label{introduction}

Nowadays, the amount of digital data generated every day is mind-blowing, while the pace of data generation is still accelerating. To deal with such amount of information, plenty of automatic solutions have been proposed and applied by various research communities such as the database, data mining and computer vision. Meanwhile, crowdsourcing has been adopted as a key problem-solving approach to information collection to address problems difficult for computers, particularly the ground truth label/score collection for the training of supervised machine learning models. There are many crowdsourcing platforms, such as Amazon Mechanical Turk (MTurk) \footnote{https://www.mturk.com/} and Crowd Flower \footnote{http://crowdflower.com/} that have been widely used \cite{li2016crowdsourced}.


Crowdsourcing tasks are mostly performed in an uncontrolled environment. The lack of control on many factors such as the people (i.e. the workers/annotators), the procedure, and the environment introduces a considerable amount of noise and unreliability, leading to less trusted results than that produced from the better-controlled test environment. Such low-quality answers further make the inference of correct answers (i.e. the so-called ground truth) a challenging task. Some existing crowdsourcing paradigms rely on redundancy-based methods to discover annotator's quality, such as using a ``golden task" before the test or a hidden task during the test \cite{zheng2017truth}. However, the lack of capability to cope with the uncertainty of annotator's reliability and crowdsourcing task difficulty largely limits the efficiency of such approaches.

Generally, we believe that the noise in labeling task mainly comes from two aspects, i.e., annotator's reliability and task's difficulty. We use \emph{reliability} to measure how likely an annotator will respond to a question seriously \cite{ye2017probabilistic}. In one labeling task, the reliable annotators answer the question seriously, while the unreliable annotators answer the question either by picking a random answer or the same answer (arbitrarily for the payment), or even maliciously giving false answers to trick the system. Annotator's reliability may vary during the labeling procedure, i.e., some annotators always give wrong answers, but some annotators may give wrong answers only a few times. In addition, we should notice that for some tasks, different annotators may have different opinions. Generally, most of the reliable annotators agree with the population's consensus whereas some of the annotators do have their own different opinions which we should respect. Their answers should be thus considered as divergent instead of unreliable or untrustworthy. The other aspect to be considered is the task \emph{difficulty}, which is the major source of noise as well. In a more difficult task, it is more likely that the annotators would give a wrong answer. The level of task difficulty determines the probability to obtain noise data. In conclusion, to infer the ground truth label, the noise from annotator and task difficulty should be considered, modeled and removed.


The traditional methods, such as Majority Voting, Mean or Median would fail to resolve this issue as they regard every annotator with the same reliability and every task with the same level of difficulty. In the database and data mining research communities, various models have been proposed \cite{zheng2017truth}\cite{li2016survey}\cite{ma2015faitcrowd}. Nevertheless, most of the models are designed for a particular application, either for the discrete labeling case, or for continuous labeling case. In addition, their modeling on annotator's behavior cannot handle different unreliable behaviors such as random/repeated/malicious labeling. Furthermore, when considering a single value (discrete or continuous) as the ground truth, the model could not correctly deal with the reliable but ``different" opinion. Thus, in our study, we propose to use the categorical distribution as the ground truth of the label rather than one value to make our model applicable on different applications (class label, continuous label, decision label) as well as capture the reliable but different answers. In addition, annotator's behavior is modeled by a latent variable (probability) to switch between reliable and unreliable behavior. The underlying ground truth distribution and annotator's behavior are estimated through Maximum Likelihood Estimates (MLE) using Expectation-Maximization (EM) algorithm. The proposed model is tested on both simulated data and real-world data, where it always shows superior performance compared to other state-of-the-art methods. 

In conclusion, the contribution of our work is four-fold:

\begin{itemize}
\item{ A simple generic probabilistic model is proposed and validated for different labeling applications by considering the categorical distribution as the ground truth.}
\item{ Different opinions from reliable annotators are considered in our model, which are integrated into the ground truth distribution.}
\item{ Task difficulty can be obtained by the entropy of the ground truth distribution, which can be used for further active labeling (assigning more annotators on difficult task rather than easy task).}
\item{ Annotator's different unreliable behaviors (random, repeated, malicious etc.) can be captured and removed from the ground truth recovering. In addition, the estimated reliability level can be further used for the selection of annotators in crowdsourcing. }
\end{itemize}

The rest of this paper is organized as follows. The related work is introduced in section \ref{related_work}. Section \ref{proposed_model} describes our proposed model in detail. To validate our model, section \ref{experiment} introduces all the experiments that have been conducted. Finally, section \ref{conclusion} concludes this work. The code of this work can be found at github\footnote{https://github.com/jingnantes/AnnotationModel. This work was done when Jing Li was with University of Nantes.}.

\section{Related work}
\label{related_work}

Data labeling task can be classified into class labeling, decision (binary) labeling and continuous labeling. Generally, the annotation model  (or truth inference model) is designed for one particular type of labeling, either discrete or continuous. Among these models, we could further classify them based on how they model the task, and how they model the annotator's quality. Zheng et al. \cite{zheng2017truth} provides a very thorough and nice review on the state-of-the-art annotation models. In this section, we select the most representatives for readers' reference. 

\subsection{Class labeling}
A class labeling task is to ask annotators to select a single or multiple classes (or categories) out of the candidate classes (or categories). For example, in Computer Vision (CV) image classification tasks, the annotators are asked to label the object (cat, dog, bird, etc.) in an image where the truth is unique. In Natural Language Processing (NLP) text classification tasks, the annotators are asked to label the topics of a document where the answers are multiple. The observed labels and the ground truth labels are treated as (discrete) class labels, no ordering is concerned. 

Dawid-Skene model \cite{dawid1979maximum} was proposed in 1979 which is a classic, efficient and effective class labeling model and has been validated by many studies \cite{lakshminarayanan2013inferring}\cite{zheng2017truth}. This model uses a confusion matrix to model an annotator's quality for answering the single-choice tasks. In each confusion matrix, the index of the row represents the ground truth label, the column value represents the probability of annotator gives the column index as the answer. \cite{kim2012bayesian}\cite{venanzi2014community} and \cite{raykar2010learning} adopted the similar idea in their models.

Another typical way to model annotator's behavior is by a single quality value, which represents the ability that this annotator correctly answers a task \cite{demartini2012zencrowd}\cite{karger2011iterative}\cite{liu2012variational}\cite{aydin2014crowdsourcing}\cite{li2014resolving}. A typical representative among these models is GLAD proposed in \cite{whitehill2009whose}. In this model, the annotator's quality is in a wide range $(-\infty, +\infty)$ where the quality value $<0$ implies an adversarial annotator (who always give adversarial answers). In addition, this model considers task difficulty while Dawid-Skene model does not. When the task difficulty increases ($+\infty$), the probability of obtaining the correct answer decreases towards 0.5 rather than $1/N$ ($N$ is the total number of classes), which is not the case in real application and has been challenged by some researchers in the community \cite{lakshminarayanan2013inferring}.  

Several studies have been conducted to compare the performances of Dawid-Skene model and GLAD model on different crowdsourcing database \cite{lakshminarayanan2013inferring}\cite{zheng2017truth}. Generally, Dawid-Skene is more reliable and robust than GLAD. 

\subsection{Decision labeling}
A decision labeling task requires the annotators to provide a ``decision", \textit{True} or \textit{False}, as the answer. Class labeling task can easily be converted to several decision labeling tasks by asking, for example, ``Is this a dog in the image?" ``Is this a cat in the image?" etc. Thus, most of the class labeling models can be easily extended to the decision labeling tasks, and vice versa \cite{demartini2012zencrowd}\cite{whitehill2009whose}\cite{dawid1979maximum}\cite{zhou2012learning}\cite{kim2012bayesian}\cite{venanzi2014community}\cite{raykar2010learning}.

Similar with the class labeling model, the annotator's behavior is generally modeled by either a quality value, for example, quality values are ranged between 0 and 1, where 1 represents experts, 0.5 represents spammers, and less than 0.5 represents adversaries \cite{karger2011iterative}\cite{liu2012variational}, or a confusion matrix \cite{liu2012variational}\cite{kim2012bayesian}. The annotation procedure is generally simulated by a Bernoulli distribution. 

Different from the models mentioned above, \cite{yan2010modeling} is a supervised learning model. The annotation procedure is modeled by a Gaussian distribution, where the mean is determined by the ground truth, the variance is a function of the annotator's expertise and task difficulty represented by the features of the input signal. As this model is supervised learning, the features of the input signal need to be trained before the test. Nevertheless, in real applications, such kind of supervised learning procedure is generally not applicable.

\subsection{Continuous labeling}

The continuous labeling task requires the annotators to provide a numerical value as the answer. This value can be ordinal, or continuous. For example, in a movie review website, the users are asked to provide their opinion of this movie on a Likert scale from 1 to 5, where 1 represents bad and 5 represents excellent. In a video quality assessment experiment, the observers are asked to rate the quality of the video on a scale of [0, 100], the obtained score can be any continuous value in this range. Generally, in continuous labeling, the underlying ground truth is considered as a continuous real value \cite{raykar2010learning}\cite{rogers2010semi}\cite{lakshminarayanan2013inferring}\cite{li2017recover}. If the observed value is an ordinal discrete value, there is a latent threshold for clipping the underlying continuous ground truth to the observed ordinal value. This threshold is either determined by the object characteristics \cite{rogers2010semi}, or by annotators \cite{raykar2012annotation}, or just artificially decided \cite{lakshminarayanan2013inferring}.  

A very first continuous annotation model is proposed in \cite{raykar2010learning} where the observed label is generated from a Gaussian distribution, with the unknown ground truth as the mean, and the accuracy of the annotator as the precision (inverse of the variance). This model is further developed by \cite{lakshminarayanan2013inferring} with the similar recipe, except for that the task difficulty is included in the model, together with the annotator's expertise as a multiplier term of the variance. In addition, Gamma priors on both terms of the variance have been imposed. Furthermore, a latent variable that describes the probability that an annotator switches between normal behavior and abnormal (spammer) behavior is used in the model, which is a pioneer work at that time. 

Different from the above, Li et.al \cite{li2017recover} propose to model the annotator's behavior by annotator's bias and annotator's inconsistency. Annotator's bias captures the effect that an annotator always underestimates or overestimates the truth of the task while annotator's inconsistency captures the variance of the attentiveness of the annotator on labeling. In addition, the task difficulty is also considered in the model. Overall, the annotation procedure is modeled by a Gaussian distribution, with ground truth continuous score plus annotator's bias as the mean, and task difficulty plus annotator's inconsistency as the variance term. However, this model may fail to model a spammer's behavior as it cannot be expressed by a Gaussian.


\subsection{Other labeling}
In addition to the above mentioned labeling, there are also some other types of labeling. For example, in a questionnaire of evaluating video quality \cite{strohmeier2010open}, the annotators are asked to use their own vocabulary to describe their perceptual experience of this video. In \cite{callison2009fast}, the annotators are asked to translate 10 sentences (with their own language, for example, French, German, Spanish...) to English. Another task in \cite{callison2009fast} is to ask the annotators to answer some reading comprehension questions. These labeling tasks are generally more open, and thus more difficult to handle. In our work, we only focus on class labeling, decision labeling and continuous ordinal labeling.

\section{Proposed annotation model}
\label{proposed_model}

In this section, we firstly describe the scope of the problems that our model can be applied. Then, the proposed model is introduced in detail. The parameters updates for the ground truth distribution as well as the latent annotator's behavior are provided, which are based on MLE using EM algorithm. Finally, the application of our model on different task scenarios are introduced. For simplicity of the explanation, all the notations are summarized in Table \ref{Tab:notation}.

\begin{table*}
  \caption{Notation}
  \label{Tab:notation}
  \begin{tabular}{ll}
    \toprule
   Notation & Description \\
\midrule
$S$   & the total number of annotators\\
$E$ & the total number of test objects\\
$N$    & the total number of candidate labels (categories)        \\
$L_n$    & the $n_{th}$ label, $n$ = 1,2,...,$N$ \\
$n$ & the $n_{th}$ label (the abbreviation of $L_n$)\\
$y_{e,s}$ & the label given by annotator $s$ for object $e$ according to the underlying ground truth distribution\\
$x_{e,s}$ & the label given by annotator $s$ for object $e$ according to annotator's irregular behaviors\\
$r_{e,s}$   & the label provided by annotator $s$ to object $e$\\
$z_{e,s}$ & latent variable which follows Bernoulli distribution determined by annotator $s$\\
$Cat(y|\mathbf{\theta_e})$ & the ground truth categorical distribution for object $e$\\
$\theta_{e,n}$      & the probability of obtaining label $n$ in one trial for object $e$\\
$\epsilon_s$     & the probability that annotator $s$ gives the label seriously       \\
$\mathbf{\pi_s}$   & the irregular behavior of annotator $s$ \\
$\delta$ & estimated parameters, $\delta = (\mathbf{\theta}, \mathbf{\epsilon}, \mathbf{\pi})$\\
$A$ & the set of all labeled objects and all annotators who have labeled\\
$l_{e,i}$ &  the set of annotators who labeled the object $e$ with label $i$.\\
$l_s$ & the set of objects labeled by annotator $s$ \\
$l_e$ &  the set of annotators who labeled object $e$\\
$\mu_{e,s,r_{e,s}}$ & for annotator $s$, object $e$, the probability that observed label $r_{e,s}$ comes from ground truth categorical distribution\\
$\lambda_e, \lambda_s$ & two Lagrange multipliers in EM algorithm \\
$Q$ & log likelihood in EM algorithm\\
$thr$ & convergence threshold for EM algorithm  \\
$v$ & the ground truth value (can be continuous or discrete) \\
    \bottomrule
  \end{tabular}
\end{table*}

\subsection{Problem setup}
\label{prob_setup}

Assuming that there are $S$ annotators and $E$ objects (e.g., products, images, movies, websites) to be labeled. The labels that can be chosen are $L_n, n=1,2,...,N$. $L_n$ can be an ordinal number ($1,2,3$ in Likert scale) or a category \{cat, dog, bird,...\}. The total number of labels is $N$. For ease of later mathematical expression, we use $n$ to interchangeably represent label $L_n$. Let $r_{e,s}$ denote the label provided by annotator $s$ to object $e$. The underlying ground truth label for object $e$ is a categorical distribution $Cat(y|\mathbf{\theta_e} ) = \prod_{n=1}^{N}\theta_{e,n}^{[y=n]}$, $\theta_{e,n}$ is the probability of obtaining label $n$ in one trial for object $e$, $\sum_{n=1}^{N}\theta_{e,n} = 1$.  $\left [ y=n\right]$ equals to 1 if $y=n$. This assumption allows us to adapt this model to different applications such as class labeling, decision labeling or continuous ordinal labeling. It should be noted that unlike the NLP document classification problem (where each document may belong to different classes), in our model, each annotator can only choose one label from the candidates. In addition, our model cannot be applied to the case that the obtained label is a continuous numeric value. 

The motivation of using a categorical distribution is that, for example, in a product review application, different reviewers' opinions cannot be the same, which are subject to their feeling and expectation, thus, the ground truth of the judgment of a product should be described by a distribution rather than a score. To get a general consensus idea from population, expectation can thus be calculated and used. Another typical example is the computer vision object classification problem, where the ground truth label is a fixed class. For object $e$, if the ground truth label is $v$, we have $\theta_{e,v}$ = 1, for others, $\theta_{e,n} $ = 0, $n\neq v$. In this case, the ground truth is a special case of the categorical distribution. Furthermore, the categorical distribution can also be applied to the decision labeling tasks where the ground truth follows a Bernoulli distribution, another special case of the categorical distribution.


In most of the existing studies, it is assumed that the label provided by an annotator is affected by the task difficulty and the annotator's quality. Any different opinion from the ground truth (a single value) is considered as an error. This is not true in some applications where there is no single-value ground truth and where we respect everyone's serious different opinion. In our model, task difficulty and observer's different opinion are integrally described by the categorical distribution. Naturally, if not specified, we consider that the annotators in a task is a representative of the population, thus, the obtained distribution reflects the global people's opinion, where different opinions exist. But still, this model can be applied to annotator groups with the similar expertise, e.g., experts, people with a specific profession, or a specific gender to infer the underlying behavior of that group.  

\subsection{Distribution-Behavior model}
\label{sec:distribution_behavior_model}

As we assumed in the previous section, the underlying ground truth for an object is a categorical distribution. In an observation, the label given by an annotator is determined by the underlying ground truth distribution as well as the annotator's behavior. Similar to \cite{lakshminarayanan2013inferring}, in our model, we consider that each annotator has a probability to provide a wrong answer, we call it ``irregular" answer. In \cite{kara2015modeling}, the authors classify annotator's behavior into eight categories, i.e., competent, spammers, adversaries, positively biased, negatively biased, unary annotators, binary annotators, and ternary annotators. In our study, we reduce the number of irregular behaviors into four categories, which can still cover the ones described in \cite{kara2015modeling}. They are: 
\begin{itemize}
\item{\textbf{Random Label}: The annotator always randomly select a label from $1$ to $N$, which follows a uniform distribution.}
\item{\textbf{Repeated Label}: This is also called ``position bias" \cite{blunch1984position}, which is to model the annotator's behavior that he/she always select the same label no matter what objects are provided.}
\item{\textbf{Inverted Label}: It means that the annotator may misunderstand the task, or intentionally to give an inverted label than the true label he/she should provide, i.e., he/she is an adversarial annotator.}
\item{\textbf{Mixed Label}: This is used to model the other irregular behaviors, which can be considered as a random combination of all previously mentioned ones.}
\end{itemize}

The probability that an annotator $s$ gives an irregular answer is $1-\epsilon_s$, where $\epsilon_s$ represents the reliability of this annotator. In the whole labeling procedure, we consider an annotator whose $\epsilon_s < 0.5$ as a ``spammer". 

The graphical model of our proposed distribution-behavior annotation model is shown in Figure \ref{fig:graph_model}. In one trial, the provided label $r_{e,s}$ is drawn from two mixture models, one is the ground truth categorical distribution $Cat(y|\mathbf{\theta_e})$, the other is the annotator's irregular behavior modeled by a discrete distribution $D(x|\mathbf{\pi_s} ) = \prod_{n=1}^{N}\pi_{s,n}^{[x=n]}$, where $\sum_{n=1}^{N}\pi_{s,n} = 1$. The switch of the two models is determined by a latent variable $z_{e,s}$, which follows a Bernoulli distribution, i.e., $B(z_{e,s}|\epsilon_s) = \epsilon_s^{z_{e,s}}(1-\epsilon_s)^{1-z_{e,s}}, z_{e,s}\in \{0,1\}$. When the latent variable $z_{e,s}=1$, the annotator labels the object according to the underlying ground truth, otherwise, the annotator labels it ``irregularly" based on his own irregular behavior $\pi_{s}$. 

\begin{figure}
\centering
\includegraphics[width=0.8\linewidth]{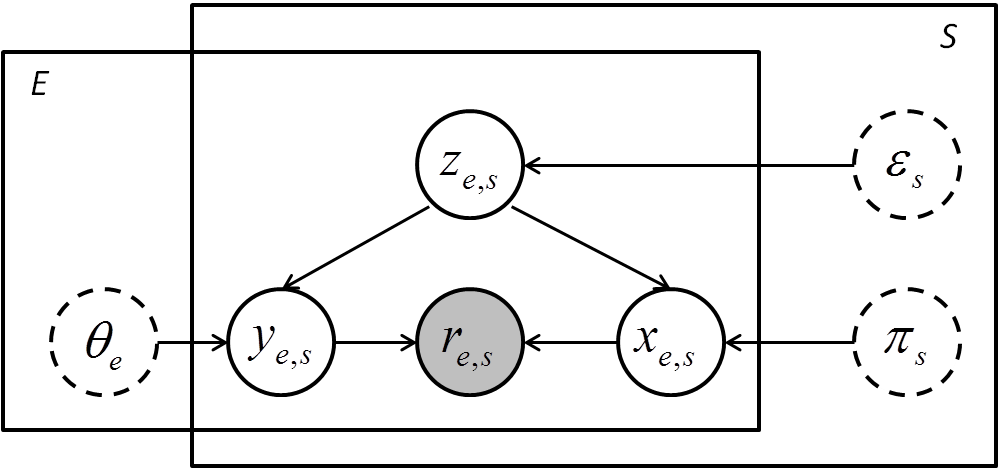}
\caption{Graphic model for our proposed distribution-behavior model. $\theta_e$, $\epsilon_s$ and $\pi_s$ are parameters, $y_{e,s}$, $x_{e,s}$ and $z_{e,s}$ are latent variables, $r_{e,s}$ is the provided label by annotator $s$.}
\label{fig:graph_model}
\vspace{-10pt}
\end{figure}

The complete conditional density is given below:

\begin{equation}
\begin{aligned}
p(Z|\epsilon) &=\prod_{e,s\in A}B(z_{e,s}|\epsilon_s)\\
p(X|\pi) &= \prod_{e,s\in A}D(x_{e,s}|\mathbf{\pi_s})\\
p(Y|\theta) &= \prod_{e,s\in A}Cat(y_{e,s}|\mathbf{\theta_e})\\
p(R|X,Y,Z) &= \prod_{e,s\in A}p(x_{e,s})^{[z_{e,s}=0]}p(y_{e,s})^{[z_{e,s}=1]}\\
\end{aligned}
\end{equation}

\noindent where $A$ represents the set of all labeled objects and all annotators. Thus, we have:
\begin{equation}
\begin{aligned}
\centering
p(R|Y,X,Z, \pi, \epsilon, \theta) &=
\sum_{Z}p(R|X,Y,Z)p(Z|\epsilon)\\
&= p(Z=1|\epsilon)p(Y|\theta)\\
&+p(Z=0|\epsilon)p(X|\pi)\\
&=\prod_{e,s\in A}[\epsilon_s(\prod_{n=1}^{N}\theta_{e,n}^{[r_{e,s}=n]})\\
&+(1-\epsilon_s)(\prod_{n=1}^{N}\pi_{s,n}^{[r_{e,s}=n]})]
\end{aligned}
\label{eq:likelihood}
\end{equation}
subject to:
\begin{equation}
\label{constraints}
    \begin{aligned}
    1 \geq \theta_{e,n}\geq 0, \sum_{n=1}^{N}\theta_{e,n}=1\\
1 \geq \pi_{s,n}\geq 0, \sum_{n=1}^{N}\pi_{s,n}=1\\
    \end{aligned}
\end{equation}

The objective of our model is to infer the underlying distribution, i.e., the parameters $\mathbf{\theta_{e}}$ and the annotator's reliability $\epsilon_s$. The discrete distribution to capture annotator's irregular behavior $D(X|\mathbf{\pi_s})$ will be discussed in section \ref{discover_annotator_behavior}.

\subsection{Parameter Estimation using EM algorithm}
\label{parameter_estimation}
The likelihood function can be calculated according to Equation (\ref{eq:likelihood}), thus, the parameters can be estimated by using MLE, i.e., 
\begin{equation}
\label{eq:likelihood2}
    \hat{\delta} =  \arg \underset{\delta}{max}\log p(R|\delta)
\end{equation}
$\delta = (\theta, \epsilon, \pi)$ are the parameters.

Since there is a latent variable in our model, there is no analytic solution for the parameters. In this paper, the EM algorithm is utilized. $\delta^{(0)}= (\theta^{(0)}, \epsilon^{(0)}, \pi^{(0)})$ denotes the initialized parameters, and $\delta^{(i)}$ = ($\theta^{(i)}$, $\epsilon^{(i)}$, $\pi^{(i)})$ denotes the parameters in the $i$-th iteration. The whole EM procedure is provided as follows (the source code will be available in Github).

\textbf{Initialization:} EM algorithm is very sensitive to initialization values. In our model, we use the following strategy. $\epsilon_s^{(0)}=0.5$, supposing that all the annotators are on the threshold of being a spammer. $\theta_{e,n}^{(0)} = \frac{\left | l_{e,n} \right |}{\sum_{k=1}^{N}\left | l_{e,k} \right |}$, $\pi_{s,n}^{(0)} = 1/N$, $n = 1, 2, ..., N$. $e$ is the index of tested object. $l_{e,n}$ denotes the set of annotators who labeled the object $e$ with label $n$. 

\textbf{E-Step:} Supposing the current estimates of parameters is $\delta^{(i)}$, for the $i+1$ iteration of E-step, calculate:
\begin{equation}
\begin{aligned}
Q(\delta, \delta^{(i)}) &= E_Z[\log p(R,Z|\delta)|R,\delta^{(i)}]\\
    &=\sum_{Z}\log p(R,Z|\delta)p(Z|R,\delta^{(i)})\\
   &=\sum_{e,s\in A} \mu_{e,s,r_{e,s}}^{(i+1)}\left [ \log\epsilon_s+\sum_{n=1}^{N}\log (\theta_{e,n}^{[r_{e,s}=n]}) \right ] \\
&+(1-\mu_{e,s,r_{e,s}}^{(i+1)})\left [\log (1-\epsilon_s)+ \sum_{n=1}^{N}\log (\pi_{s,n}^{[r_{e,s}=n]}) \right ]
\end{aligned}
\label{eq:Qfunction}
\end{equation}
where $\mu_{e,s,r_{e,s}}^{(i+1)}$ is the probability that the provided label $r_{e,s}$ comes from the ground truth categorical distribution under the current parameters $\delta^{(i)}$:
\begin{equation}
\begin{aligned}
    \mu_{e,s,r_{e,s}}^{(i+1)} &=\frac{c1}{c1+c2}\\ c1 &=\sum_{n=1}^{N}\theta_{e,n}^{(i)[r_{e,s}=n]}\\
    c2&=(1-\epsilon_s^{(i)})(\sum_{n=1}^{N}\pi_{s,n}^{(i)[r_{e,s}=n]})
\end{aligned}
\end{equation}

\textbf{M-Step:} Find out the $\delta$ that maximizes $Q(\delta,\delta^{(i)})$ as the estimates of the $i+1 th$ iterations, i.e., 
\begin{equation}
\begin{aligned}
    \delta^{(i+1)} = \arg \underset{\delta}{max}Q(\delta, \delta^{(i)})\\
    s.t. \sum_{n=1}^{N}\theta_{e,n}=1, \sum_{n=1}^{N}\pi_{s,n}=1
\end{aligned}
\end{equation}
Lagrange multipliers $\lambda_{e}$ and $\lambda_{s}$ are thus introduced for $\theta$ and $\pi$, independently. 
For completeness, we provide the whole parameter updates here.
\begin{equation}
    \begin{aligned}
    \epsilon_s^{(i+1)} &= \frac{\sum_{e\in l_s}\mu_{e,s,r_{e,s}}^{(i+1)}}{|l_s|}\\
    \theta_{e,n}^{(i+1)} &= - \frac{\sum_{s\in l_{e,n}}\mu_{e,s,n}^{(i+1)}}{\lambda_{e}}\\
    \lambda_{e}^{(i+1)}&=-\sum_{s\in l_e}\sum_{n=1}^{N}\mu_{e,s,n}^{(i+1)}\\
    \pi_{s,n}^{(i+1)} &= - \frac{\sum_{e\in l_{s,n}}(1-\mu_{e,s,n}^{(i+1)})}{\lambda_{s}}\\
    \lambda_{s}^{(i+1)}&=-\sum_{e\in l_s}\sum_{n=1}^{N}(1-\mu_{e,s,n}^{(i+1)})
    \end{aligned}
\label{eq:updates}
\end{equation}
where $l_s$ denotes the set of objects labeled by annotator $s$, $l_e$ denotes the set of annotators who labeled object $e$, $l_{e,n}$ denotes the set of annotators who labeled object $e$ with $n$, $l_{s,n}$ denotes the set of objects labeled by annotator $s$ with $n$. 

\textbf{Convergence criterion:} Evaluate the log likelihood (i.e., Q function defined in Equation \ref{eq:Qfunction}), repeat E-step and M-step until the convergence criterion is satisfied:
\begin{equation}
    \left \| Q(\delta^{(i+1)}, \delta^{(i)})-Q(\delta^{(i)},\delta^{(i)}) \right \|< thr
\end{equation}
\noindent In our model, we set $thr = 0.0001$.

\subsection{Discovering the annotator's irregular behavior}
\label{discover_annotator_behavior}
An ideal discrete distribution $D(x|\pi)$ should be able to capture annotator's diverse irregular behaviors including random label, repeated label, inverted label and other more complicated conditions as we described in the beginning of Section \ref{sec:distribution_behavior_model}. However, in reality, it is hard to find this ideal candidate. Alternatively, we consider the uniform distribution as a loose assumption on prior, which also makes the updates much easier in this model, i.e., by setting all $\pi_{s,n}=1/N, n=1, 2, ..., N$. The feasibility of the utilization of uniform distribution to capture the irregular behaviors is validated and shown in Section \ref{exp:malicious annotator}. We keep the general updates for all parameters as shown in Equation (\ref{eq:updates}) to allow for further investigation of $D(x|\pi)$ for the readers.



\subsection{Prediction of ground truth}
\label{prediction_gt}
In our model, the ground truth is a categorical distribution with parameters $\theta_{e} = (\theta_{e,1},..., \theta_{e,N})$. This model can be applied directly to the condition that there is possibility that people have different opinions on the labels of the object. In addition, our model still allows us to extend it to other applications. We will provide more details below to show how to apply our model on them.

\textbf{Continuous case:}
In the case where the underlying ground truth is by nature a continuous value, whereas the required label is an ordinal value, the ground truth could be obtained by calculation of the expectation of the estimated ordinal categorical distribution, i.e., $\hat{v}_e = \sum_{n=1}^{N}n\cdot\theta_{e,n}$.

\textbf{Discrete case:}
Both the class labeling and decision labeling belong to this discrete case. A typical class labeling problem can be considered as a special case of categorical distribution, where the correct class with probability of 1, and others with 0. A decision labeling problem can be considered as a Bernoulli distribution. Thus, in discrete labeling, our model can estimate the ground truth by calculating the mode of the predicted distribution, i.e., $\hat{v}_e = n$ such that $\theta_{e,n} = max(\theta_{e,1}, \theta_{e,2},...,\theta_{e,N})$. 


\section{Experiment}
\label{experiment}
We tested our model on two types of data sets. One is simulated data, the other is real world data. Details are shown below.

\subsection{Evaluation metrics}
\label{evaluation_metric}
Firstly, we summarize different evaluation metrics for the performance of annotation models in this part. 

\subsubsection{Classification accuracy $C_o$}
The ratio of correctly classified objects to the total number of objects:
\begin{equation}
C_o = \sum_{e=1}^{E}\left [\hat{v}_e=v_e\right ]/E
\end{equation}

\subsubsection{$F_1$ score}
A measure of a classifier's accuracy, which is the harmonic mean of precision and recall \cite{sorensen1948method}:
\begin{equation}
F_1 = 2\cdot\frac{precision \cdot recall}{precision + recall}
\end{equation}

\subsubsection{Pearson Linear Correlation Coefficient (PLCC)}
A measure of the linear correlation between two variables \cite{pearson1895note}:
\begin{equation}
PLCC = \frac{1}{E-1}\frac{\sum_{e=1}^{E}(v_e-\overline{v})(\hat{v}_e-\overline{\hat{v}})}{\sqrt{Var(v)Var(\hat{v})}}
\end{equation}
$Var(v)$ is the variance of true values $v, v=(v_1,v_2,...,v_E)$, the same applied to $Var(\hat{v})$.

\subsubsection{Spearman Rank Order Correlation Coefficient (SROCC)}
A nonparametric measure of rank correlation. Let $r = (r_1, r_2,..., r_E)$ be the rank of $v$, and $\hat{r}$ be the rank of $\hat{v}$, SROCC is calculated by:
\begin{equation}
SROCC = \frac{1}{E-1}\frac{\sum_{e=1}^{E}(r_e-\overline{r})(\hat{r}_e-\overline{\hat{r}})}{\sqrt{Var(r)Var(\hat{r})}}
\end{equation}

\subsubsection{Root Mean Square Error (RMSE)} 
A measure of how spread out the prediction errors are:
\begin{equation}
RMSE = \sqrt{\frac{\sum_{e=1}^{E}(v_e-\hat{v}_e)^2}{E}}
\end{equation}

\subsubsection{Hellinger distance}
A measure of the similarity between two probability distributions: 
\begin{equation}
H(\theta_e,\hat{\theta}_e) = \frac{1}{\sqrt{2}}\sqrt{\sum_{n=1}^{N}(\sqrt{\theta_{e,n}}-\sqrt{\hat{\theta}_{e,n}})}
\end{equation}

Higher values of $C_o$, $F_1$, PLCC, SROCC, Hellinger distance, and lower values of RMSE indicate better performance.

\subsection{Simulated data}

\subsubsection{Exp1-a: Detection of irregular annotations}
\label{exp:malicious annotator}
The objective of this experiment is to see whether or not the uniform distribution can capture different types of irregular behaviors. In this experiment, we simulate the ground truth as a categorical distribution with $N$ = 5. The categorical distribution is generated based on a Beta distribution $Be(\alpha,\beta)$, where $\alpha$ and $\beta$ are randomly selected from 1 to 10. The obtained Beta distribution is then re-scaled and clipped to form a categorical distribution.   

We simulate in total 150 objects which are labeled by 25 annotators on average. The $\epsilon_s$ is randomly selected from 0 to 1. Meanwhile, we set 20\% of the annotators $\epsilon_s$ values lower than 0.5, we call them ``spammers". 20\% is considered as spamminess ratio. 

Four types of irregular behaviors are considered. They are ``random", ``repeated", ``inverted", and ``mixed" labeling, which are generated in the following way:
\begin{itemize}
    \item \textbf{Random}: the annotator's label is randomly sampled from a uniform distribution $U(1,N)$. The observed label is a discrete label.
    \item \textbf{Repeated}: each annotator is assigned with a fixed position bias in the simulation, which is randomly sampled from a uniform distribution $U(1,N)$. Then, for a particular annotator, his/her repeated label is always the one assigned to him/her.
    \item \textbf{Inverted}: the provided label by an annotator is $N-y_{e,s}+1$. $y_{e,s}$ is the observed label according to the ground truth distribution. For example, an adversarial annotator observes ``Excellent (5)" but he/she provides ``Very Bad (1)" in the task. This is particularly applicable for decision labeling and continuous (ordinal) labeling.
    \item \textbf{Mixed}: a combination of the behaviors above in a random way. 
\end{itemize}

To obtain statistical results, each type of behavior is conducted 100 times. The evaluation methods for the performance of detection of irregular annotations are $F_1$ score for the classification of spammers (whose $\epsilon_s < 0.5$), PLCC, SROCC and RMSE between ground truth reliability level $\epsilon = \{\epsilon_1,\epsilon_2,...,\epsilon_S\}$ and the estimated values. Results are shown in Table \ref{tab:exp1a}.

\begin{table}
  \caption{Exp1-a: Performance on detection of different irregular behaviors. The values are the mean of 100 test results. $\uparrow$ means the higher the value, the better the performance. $\downarrow$ means the opposite.}
  \label{tab:exp1a}
  \begin{tabular}{lcccc}
    \toprule
Behavior type    & $F_1$ $\uparrow$ & PLCC $\uparrow$ & SROCC $\uparrow$ & RMSE $\downarrow$ \\
\midrule
Random & 0.9949 & 0.9228 & 0.9011 & 0.2071\\
Repeated & 0.9121 & 0.5835 & 0.5398 & 0.3388\\
Inverted & 0.9458 & 0.7297 & 0.7567 & 0.3296\\
Mixed & 0.9335 & 0.6922 & 0.7305 & 0.3539\\
    \bottomrule
  \end{tabular}
\end{table}

The results indicate that the uniform distribution could capture the ``random" behavior effectively, in terms of classification accuracy ($F_1$) as well as scale prediction accuracy (PLCC, SROCC and RMSE). However, for other types of behaviors, the accuracy of predicting absolute $\epsilon_s$ values is not satisfactory. Nevertheless, its classification performances are promising with $F_1$ scores all above 0.9. Thus, in reality, the estimated $\epsilon_s$ value can be used to classify the annotators as spammers or not.

\subsubsection{Exp1-b: Influence of the proportion of irregular annotations}
\label{exp:influence_irregular_annotation}
The objective of this experiment is to test the influence of the proportion of irregular behaviors on the prediction accuracy. In this experiment, we simulate the ground truth as a categorical distribution as we did in section \ref{exp:malicious annotator} with 150 objects and 25 annotators. The $\epsilon_s$ are randomly selected from 0 to 1. In addition, we set 5 levels of spamminess ratio. They are 5\%, 10\%, 15\%, 20\% and 25\%. The irregular behavior is fixed as ``mixed". Each test is repeated 100 times for statistical reliability.

The evaluation methods are RMSE and Hellinger distance between ground truth distribution and the estimated distribution. To make a comparison, the observed distribution is also evaluated. Results are shown in Figure \ref{fig:exp1b}. 

Figure \ref{fig:exp1b} shows that with the increase of the spamminess ratio, the prediction error is increasing as well, which is reasonable. An interesting finding is that when the spamminess ratio increases from 0.1 to 0.15, the prediction error of our proposed model is even smaller, which is not the case for the estimated distribution from observed labels. In addition, the prediction accuracy of observed data under spamminess ratio of 5\% is the same with our proposed model under spamminess ratio of 16\%, which indicates an increment of 11\% spamminess tolerance for our model. In conclusion, our proposed model is more robust to irregular annotations than directly using the observed labels, which is particularly applicable in crowd-sourcing labeling scenario. 

\begin{figure}
\centering
\begin{tabular}{cc}
\includegraphics[width = 0.47\linewidth]{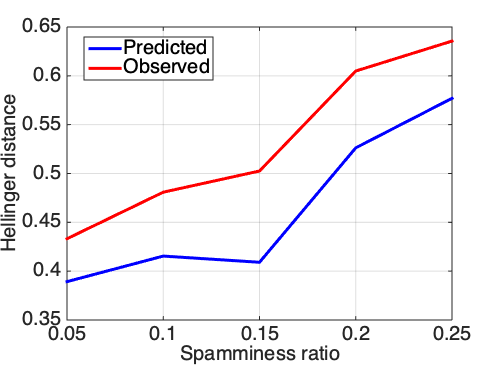} & \includegraphics[width = 0.47\linewidth]{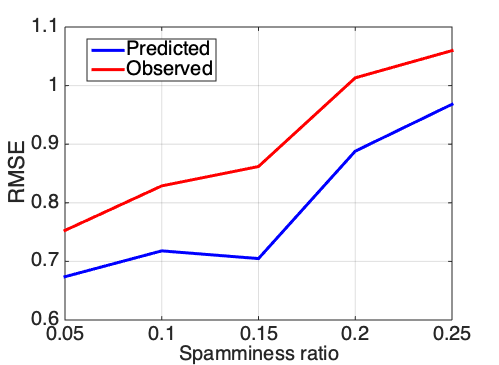}\\
\footnotesize{(a) } & \footnotesize{(b) }\\
\end{tabular}
\caption{Exp1-b: Influence of the proportion of irregular annotations on prediction. Reported values are the mean of the 100 test results. }
\label{fig:exp1b}
\end{figure}

\subsubsection{Exp1-c: Prediction accuracy}
The objective of this experiment is to study the prediction accuracy under different number of annotations. All the experimental simulation setup is similar with Exp1-b, except for that the spamminess ratio is fixed to 20\%, and we set 7 levels of annotation numbers, which are 10, 15, 20, ..., to 40. Again, for statistical reliability, each test is repeated 100 times. The evaluation methods are the same with Exp1-b. Results are shown in Figure \ref{fig:exp1c}.

\begin{figure}
\centering
\begin{tabular}{cc}
\includegraphics[width = 0.47\linewidth]{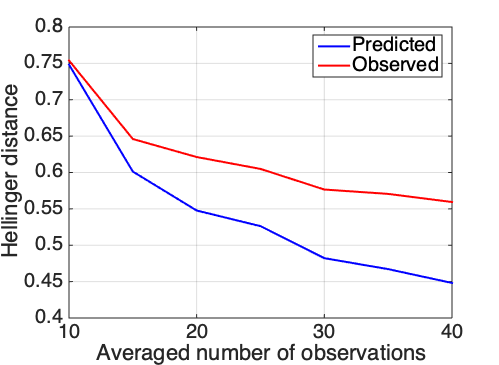} & \includegraphics[width = 0.47\linewidth]{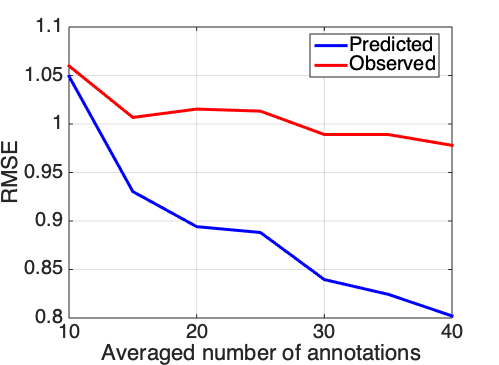}\\
\footnotesize{(a) } & \footnotesize{(b) }\\
\end{tabular}
\caption{Exp1-c: Prediction accuracy in terms of the number of annotations. Reported values are the mean of the 100 test results.}
\label{fig:exp1c}
\end{figure}

The experimental results indicate that with our model, the required number of annotations to obtain as accurate results as using observed data can be reduced significantly. An example is that to achieve the same accuracy of 40 annotations using the direct observed labels, only 20 or even fewer annotations are needed by using our proposed model.

\subsubsection{Exp1-d: Universality study}
The objective of this experiment is to test the universality of our model on other types of data rather than a categorical distribution. In this experiment, we assume that the ground truth is a continuous score for each object, which is randomly (uniformly) selected from [1, 5]. The observed score follows a Gaussian distribution where mean is the ground truth score, and the precision (inverse of the variance) is randomly sampled from a Gamma distribution $\sim Gamma(10,5)$. Clipping (round) is conducted to make the observed score as an integer and in the range of [1, 5]. There are in total 150 test objects. We set annotator's irregular behavior as ``mixed". The number of annotations is 25. Spamminess ratio is 25\%. Each test is repeated 100 times.

We compare our proposed model with the traditional methods, i.e., Majority Vote, Mean, and the state-of-the-art models which are based on Gaussian assumption, i.e., ordinal-discrete-mixture model \cite{lakshminarayanan2013inferring} and Li's MLE model \cite{li2017recover}. The evaluation methods are PLCC, SROCC and RMSE between the predicted label and ground truth. In our model, we use the expectation of the estimated distribution to predict the ground truth. Results are shown in Table \ref{tab:exp1d}.

\begin{table}
  \caption{Exp1-d: Universality study of the proposed model. The values are the mean of 100 test results. $\uparrow$ means the higher the value, the better the performance. $\downarrow$ means the opposite.}
  \label{tab:exp1d}
  \begin{tabular}{lccc}
    \toprule
Model    & PLCC$\uparrow$  & SROCC$\uparrow$ & RMSE$\downarrow$ \\
\midrule
Mean & 0.8808  & 0.8997 & 0.6021 \\
Majority & 0.8918 & 0.8805 & 0.7053 \\
Li \cite{li2017recover} & 0.9138& 0.9176& 0.5792 \\
Ord-dis-mix\cite{lakshminarayanan2013inferring} & \textbf{0.9699} & \textbf{0.9680} & \textbf{0.3050} \\
Proposed & 0.9432& 0.9453 & 0.4657 \\
    \bottomrule
  \end{tabular}
\end{table}

As the Ord-discrete-mix model is designed based on Gaussian where the annotator's irregular behavior is also considered, it is reasonable that its performance is the best. It is interesting to notice that our proposed model performs the second best, which is better than another Gaussian based model, i.e., Li's MLE model. This result validates that our model has excellent generality under different types of data assumption. Mean and Majority methods show their weak robustness ability when there is a large number of irregular labelings.

\subsection{Real-world data}
In this section, we compare our model with the state-of-the-art ground truth prediction models, i.e., Dawid-Skene (D\&S) \cite{dawid1979maximum}, GLAD \cite{whitehill2009whose}, Ord-bin \cite{raykar2009supervised}, Ord-dis-mix \cite{lakshminarayanan2013inferring} and Li's MLE model \cite{li2017recover} on real-world data sets. To verify the generality of our model in different applications, three different datasets are used. Details are shown in the following parts.

\subsubsection{Crowd Dog Classification}
\label{sec:crowd_dog}
In this experiment, the data from \cite{fang2017improving} are used. The task is to ask the annotators to recognize the breed of the dog in a given image. There are in total 250 images under test, and 4 types of dogs, i.e., $N$=4. Each image is labeled by 17 annotators on average. The data also provides ground truth labels. 

As this data is obtained from a class labeling task, discrete models, i.e., D\&S \cite{dawid1979maximum}, GLAD \cite{whitehill2009whose}, Ord-bin \cite{raykar2009supervised} are used for comparison. For our model, according to the predicted distribution, we select the category with the highest probability as the predicted label (i.e., mode). The evaluation method for the performance of the models is the classification accuracy $C_o$ and $F_1$ score. Results are shown in Table \ref{tab:crowd-dog}.

\begin{table}
  \caption{Classification performance of different models on Crowd Dog dataset\cite{fang2017improving}. $\uparrow$ means the higher the value, the better the performance. $\downarrow$ means the opposite.}
  \label{tab:crowd-dog}
  \begin{tabular}{lcccc}
    \toprule
Model    & Ord-bin \cite{raykar2009supervised} &  D\&S \cite{dawid1979maximum} & GLAD \cite{whitehill2009whose} &  Proposed \\
\midrule
$C_o$ $\uparrow$ & 0.6280  & 0.6320 & \textbf{0.6480} &  \textbf{0.6480}  \\
$F_1$ $\uparrow$ & 0.4862 & 0.5000 & \textbf{0.5368} &  0.5217  \\
    \bottomrule
  \end{tabular}
\end{table}

According to the results, Ord-bin performs the worst and GLAD performs the best. Our proposed model performs the second best, which achieves the same classification accuracy with GLAD, however, the $F_1$ score is a little bit lower. 

We also use our model to check the spamminess ratio of this dataset, which is 26.21\%. As there is no ground truth for the spammer check, this result is only used for reference.

\subsubsection{Face Emotion Identification}
\label{face}
In this experiment, we use the Face Emotion Identification dataset \footnote{https://github.com/zhydhkcws/crowd\_truth\_infer}, which comprises 5242 labels applied on 584 face images collected by 27 annotators. The task is to ask the annotators to identify the sentiment of the face in the image with the label ``neutral(0)", ``happy(1)", ``sad(2)", and ``angry(3)". The ground truth labels are also provided.  
The compared models as well as the evaluation methods are exactly the same with section \ref{sec:crowd_dog}. Results are shown in Table \ref{tab:face}.

Interestingly, the performances of the models on this dataset are a little bit different from previous one (i.e., Crowd Dog dataset). In this dataset, Ord-bin performs the best while GLAD performs the worst, which is inverse with the results of Crowd Dog dataset. Our proposed model performs the second best, and D\&S model performs the third, which keep the same with the results of Crowd Dog dataset. The results to some extent verify the conclusion from \cite{zheng2017truth} that there is generally no ``perfect" model that always outperforms the others, but generally speaking, D\&S model is a robust one. In these two datasets, we demonstrate that our proposed model is more robust than D\&S. 

In addition, according to our model, the detected spamminess ratio for this dataset is 0, which means that this data is quite reliable.

\begin{table}
  \caption{Classification performance of different models on Face emotion dataset$^{3}$. $\uparrow$ means the higher the value, the better the performance. $\downarrow$ means the opposite.}
  \label{tab:face}
  \begin{tabular}{lcccc}
    \toprule
Model    & GLAD \cite{whitehill2009whose} &  D\&S \cite{dawid1979maximum} & Ord-bin \cite{raykar2009supervised} &  Proposed \\
\midrule
$C_o$ $\uparrow$ & 0.5976 & 0.6318 & \textbf{0.6490} & 0.6370  \\
$F_1$ $\uparrow$ & 0.5155 & 0.5356 & \textbf{0.5514} & 0.5431  \\
    \bottomrule
  \end{tabular}
\end{table}

\subsubsection{Movie Review}
\label{movie_review}
In this experiment, we use the MovieLens 20M dataset\footnote{https://grouplens.org/datasets/movielens/}, which comprises 20 million ratings applied to 27,000 movies by 138,000 users on a 5-level scale. As the task is about the opinion of the users on the movie, there is no real ground truth. In our study, we select a subset of this dataset containing 5662 movies labeled by 15147 annotators as the validation dataset, ensuring that each movie is labeled by at least 30 users, and each annotator labeled at least 30 movies. The obtained mean of all annotator's rating is considered as the ground truth. In addition, to test our model, we further sample a small dataset from the validation dataset, which contains 2833 ratings with 174 movies labeled by 69 users. The ground truth ratings of the 174 movies are obtained by 1452 annotations/movie on average, while in the test data each movie is annotated by 16 users on average. 

As this experiment is a continuous case, for our model, we use the expectation of the predicted distribution as the predicted score. The compared models in this experiment are D\&S \cite{dawid1979maximum}, GLAD \cite{whitehill2009whose}, Ord-bin \cite{raykar2009supervised}, Ord-dis-mix \cite{lakshminarayanan2013inferring} and Li's MLE model \cite{li2017recover}. The evaluation methods are PLCC, SROCC and RMSE between the ground truth and the predicted score. Results are shown in Table \ref{tab:movielens}. The predicted spamminess ratio in this data is only 1\%.

According to the results, as the D\&S, GLAD, Ord-bin models are proposed for discrete labeling though they are also applicable on ordinal data, their performances are generally worse than the continuous models such as Ord-dis-mix and Li's MLE. However, our proposed model performs the best. In addition, considering that the spamminess ratio in this dataset is quite low, we may draw a conclusion that in this experiment, it is the underlying complicated data distribution that determines the performances of different models, which further demonstrates that our proposed model is a generic model applicable in different data patterns and different applications.


\begin{table}
  \caption{Performances of different models on MovieLens dataset. $\uparrow$ means the higher the value, the better the performance. $\downarrow$ means the opposite.}
  \label{tab:movielens}
  \begin{tabular}{lccc}
    \toprule
Model    & PLCC $\uparrow$ & SROCC $\uparrow$ & RMSE $\downarrow$ \\
\midrule
D\&S \cite{dawid1979maximum} & 0.4073 & 0.4957 & 1.2287  \\
GLAD \cite{whitehill2009whose} & 0.5347 & 0.6029 & 0.6361 \\
Ord-bin \cite{raykar2009supervised} & 0.7122 & 0.7166 & 0.5849 \\
Ord-dis-mix\cite{lakshminarayanan2013inferring} & 0.7066 & 0.7282 & 0.4292 \\
Li \cite{li2017recover} & 0.8369 & 0.8219 & 0.2483 \\
Proposed & \textbf{0.8620} & \textbf{0.8420} & \textbf{0.2228}\\
    \bottomrule
  \end{tabular}
\end{table}

\section{Conclusion}
\label{conclusion}
In this paper, we propose to use a categorical distribution to represent the underlying ground truth rather than a single value as other works did. The usage of a distribution allows us to 1) apply this model on any labeling tasks, no matter class labeling, decision labeling, or continuous (ordinal) labeling; 2) model the serious but different opinions from the population which we should respect. In addition, a latent variable is introduced to model the probability that an annotator may switch between a reliable and unreliable annotator at any time during a labeling task, which is often happening in real life, especially crowdsourcing scenario. Furthermore, different types of irregular behaviors, such as random labeling, repeated labeling, inverted labeling, and others can be captured by our model by simply using a uniform distribution. The proposed model has been validated on both simulated data and real-world data, where it always shows promising performances than the others in terms of prediction accuracy and robustness to irregular behaviors. In the future, we may consider to extend this work to pair comparison experiment to identify the irregular behavior, discover the personal preference, and infer the consensus opinion in population, which can be applied to recommendation system, A/B test, player matching system, etc., accordingly.




%
\bibliographystyle{IEEEtran}
\bibliography{sample-base}

%
\newpage
\appendix

\section{Parameter Estimation}
In our model, $Q$ function is defined in Equation \ref{eq:Qfunction}. For M-step, we need to find  $ \delta^{(i+1)} = \arg \underset{\delta}{max}Q(\delta, \delta^{(i)})$
subject to:
$\sum_{n=1}^{N}\theta_{e,n}=1, \sum_{n=1}^{N}\pi_{s,n}=1$. 
Thus, we have the updated $Q'$ function:
\begin{equation}
    Q' = Q + \sum_{e \in l_s}\lambda_e(\sum_{n}\theta_{e,n}-1)+\sum_{s \in l_e}\lambda_s(\sum_{n}\pi_{s,n}-1)
\end{equation}
$\lambda_e,\lambda_s$ are Lagrange multipliers. To find the maxima, we need to calculate:

\begin{equation}
 \begin{aligned}
    \frac{\partial Q'}{\partial \epsilon_s} &= \sum_{e\in l_s}\left ( \mu_{e,s,r_{e,s}}^{(i+1)}\frac{1}{\epsilon_s}+(1-\mu_{e,s,r_{e,s}}^{(i+1)})\frac{-1}{1-\epsilon_s} \right )\\
&=\frac{1}{\epsilon_s}\sum_{e\in l_s}\mu_{e,s,r_{e,s}}^{(i+1)}-\frac{1}{1-\epsilon_s}\sum_{e\in l_s}(1-\mu_{e,s,r_{e,s}}^{(i+1)})\\
&=\frac{1}{\epsilon_s}\sum_{e\in l_s}\mu_{e,s,r_{e,s}}^{(i+1)}-\frac{1}{1-\epsilon_s}(\left | l_s \right |-\sum_{e\in l_s}\mu_{e,s,r_{e,s}}^{(i+1)})\\
&=\frac{1}{\epsilon_s(1-\epsilon_s)}(\sum_{e\in l_s}\mu_{e,s,r_{e,s}}^{(i+1)}-\left | l_s \right |\epsilon_s)\\
 \end{aligned}
\end{equation}

\begin{equation}
\begin{aligned}
 \frac{\partial Q'}{\partial \theta_{e,n}} &= \sum_{s \in l_{e,n}}(\mu_{e,s,r_{e,s}}^{(i+1)}\cdot \frac{1}{\theta_{e,n}})+\lambda_e\\
 \frac{\partial Q'}{\partial \pi_{s,n}} &= \sum_{e \in l_{s,n}}(1-\mu_{e,s,r_{e,s}}^{(i+1)})\cdot \frac{1}{\pi_{s,n}}+\lambda_s\\
  \frac{\partial Q'}{\partial \lambda_e} &= \sum_{n=1}^{N}\theta_{e,n}-1 \\
   \frac{\partial Q'}{\partial \lambda_s} &= \sum_{n=1}^{N}\pi_{s,n}-1 \\
 \end{aligned}
\end{equation}

Let $\frac{\partial Q'}{\partial \epsilon_s} = 0$, $\frac{\partial Q'}{\partial \theta_{e,n}} = 0$, $\frac{\partial Q'}{\partial \pi_{s,n}} = 0$,  $\frac{\partial Q'}{\partial \lambda_e} = 0$, $\frac{\partial Q'}{\partial \lambda_s} = 0$, we have:
\begin{equation}
 \begin{aligned}
 \epsilon_s^{(i+1)} &= \frac{\sum_{e\in l_s}\mu_{e,s,r_{e,s}}^{(i+1)}}{|l_s|}\\
 \theta_{e,n}^{(i+1)} &= - \frac{\sum_{s\in l_{e,n}}\mu_{e,s,n}^{(i+1)}}{\lambda_{e}}\\
 \pi_{s,n}^{(i+1)} &= - \frac{\sum_{e\in l_{s,n}}(1-\mu_{e,s,n}^{(i+1)})}{\lambda_{s}}\\
  \lambda_{e}^{(i+1)}&=-\sum_{s\in l_e}\sum_{n=1}^{N}\mu_{e,s,n}^{(i+1)}\\
 \lambda_{s}^{(i+1)}&=-\sum_{e\in l_s}\sum_{n=1}^{N}(1-\mu_{e,s,n}^{(i+1)})
 \end{aligned}
\end{equation}


\end{document}